\documentclass[conference]{IEEEtran}
\IEEEoverridecommandlockouts

\usepackage{amsmath,amssymb,amsfonts}
\usepackage{amsthm}
\usepackage{mathtools}
\usepackage{graphicx}
\usepackage{booktabs}
\usepackage{textcomp}
\usepackage{xcolor}
\usepackage{microtype}

\usepackage{tikz}
\usepackage{pgfplots}
\pgfplotsset{compat=1.18}
\usetikzlibrary{positioning}

\usepackage[numbers,sort&compress]{natbib}
\usepackage{hyperref}
\usepackage[capitalize,noabbrev]{cleveref}

\pgfplotsset{
  every axis/.append style={
    font=\footnotesize,
    label style={font=\footnotesize},
    tick label style={font=\scriptsize},
    legend style={font=\scriptsize, draw=none, fill=none, inner sep=1pt},
    legend cell align=left,
  },
}

\def\nQuestions{40}
\def\nModels{7}
\def\creFracQuestionPct{85.2}
\def\creFracModelPct{0.5}
\def\looFracQMin{83.1}
\def\looFracQMax{90.2}
\def\rankTauBrierlThreeVsslope{0.143}
\def\rankPBrierlThreeVsslope{0.773}
\def\rankTauBrierlThreeVsdrrslope{0.429}
\def\rankPBrierlThreeVsdrrslope{0.239}
\def\rankTauBrierlThreeVsdrrres{0.714}
\def\rankPBrierlThreeVsdrrres{0.0302}
\def\nCellsOrig{268}
\def\nCellsBlind{280}
\def\nCellsIncomplete{12}
\def\nNullRecords{181}
\def\crOrigNaiveBeta{-0.531}
\def\crOrigCtrlBeta{-1.781}
\def\crBlindNaiveBeta{-0.506}
\def\crBlindNaiveLo{-0.785}
\def\crBlindNaiveHi{-0.227}
\def\crBlindCtrlBeta{-1.655}
\def\crBlindCtrlLo{-1.801}
\def\crBlindCtrlHi{-1.510}
\def\crBlindCommonBeta{-1.654}
\def\crBlindCommonLo{-1.820}
\def\crBlindCommonHi{-1.487}
\def\crOrigSlopeTwelveBeta{-0.638}
\def\crOrigSlopeTwelveLo{-0.681}
\def\crOrigSlopeTwelveHi{-0.595}
\def\crBlindSlopeTwelveBeta{-0.601}
\def\crBlindSlopeTwelveLo{-0.694}
\def\crBlindSlopeTwelveHi{-0.509}
\def\uMean{0.338}
\def\uSd{0.055}
\def\uMin{0.271}
\def\uMax{0.412}
\def\KthreeUMean{0.135}
\def\corrUH{+0.27}
\def\couplingSqrtBrier{+0.131}
\def\couplingH{+0.227}
\def\uMeanBlind{0.085}
\def\uMinBlind{0.018}
\def\uMaxBlind{0.126}
\def\corrUHBlind{+0.87}
\def\uTableRowsBoth{Haiku 4.5 & 0.577 & 0.0489 & 0.289 & 0.0030 & 0.018 \\
Opus 4.7 & 0.510 & 0.0583 & 0.298 & 0.0246 & 0.126 \\
Sonnet 4.6 & 0.523 & 0.0787 & 0.412 & 0.0218 & 0.108 \\
GPT-4o & 0.533 & 0.0655 & 0.351 & 0.0101 & 0.053 \\
GPT-5 & 0.500 & 0.0690 & 0.345 & 0.0253 & 0.126 \\
GPT-5 mini & 0.502 & 0.0799 & 0.402 & 0.0148 & 0.077 \\
o3 & 0.507 & 0.0534 & 0.271 & 0.0164 & 0.083 \\}
\def\aggOrigLZero{0.521}
\def\aggOrigLThree{0.687}
\def\aggCoordsOrig{(0,0.5212) (1,0.5226) (2,0.6539) (3,0.6870)}
\def\aggBlindLZero{0.519}
\def\aggBlindLThree{0.566}
\def\aggCoordsBlind{(0,0.5191) (1,0.5329) (2,0.5580) (3,0.5660)}
\def\rankTauHake{+0.333}
\def\rankPHake{0.381}
\def\rankTauMurphyRes{+0.619}
\def\rankPMurphyRes{0.069}
\def\cellEpsMedian{0.023}
\def\cellEpsPninety{0.192}
\def\fracMonoCells{34}
\def\corrChangeBaseline{-0.55}
\def\corrChangeOldham{-0.02}
\def\blomqvistObs{-0.582}
\def\blomqvistCorr{-0.564}
\def\sigmaRatioPct{4.1}
\def\manifoldRBrLZeroLThree{+0.73}
\def\manifoldQVarPct{87}
\def\manifoldMVarPct{0.8}
\def\primarySlopeOnSqrtBrL{-0.195}
\def\primaryRBrLZeroLThree{+0.64}
\def\primaryQVarPct{77}
\def\primaryMVarPct{3.2}
\def\acledSlopeOnSqrtBrL{+0.273}
\def\acledRBrLZeroLThree{-0.05}
\def\acledQVarPct{61}
\def\acledMVarPct{7.5}
\def\brierLThreeGptFourOFull{0.140553}
\def\brierLThreeHaikuFull{0.140631}

\pgfplotsset{reasoning/.style={}, nonreasoning/.style={}}

\theoremstyle{plain}
\newtheorem{proposition}{Proposition}

\theoremstyle{remark}

\usepackage{eso-pic}

\begin{document}

\title{Update Slope Is Bounded by Prior Calibration:\\
       A Structural Bound and Its Contingent Empirical Signature}

\author{\IEEEauthorblockN{Ben Jenkins}
\IEEEauthorblockA{\textit{Dept.\ of Electrical Engineering and Computer Science}\\
\textit{Florida Atlantic University}\\
Boca Raton, FL, USA\\
benjaminjenk2024@fau.edu}
\and
\IEEEauthorblockN{Mihaela Cardei}
\IEEEauthorblockA{\textit{Dept.\ of Electrical Engineering and Computer Science}\\
\textit{Florida Atlantic University}\\
Boca Raton, FL, USA\\
mcardei@fau.edu}}

\maketitle
\AddToShipoutPictureFG*{\AtPageLowerLeft{\put(\LenToUnit{0.5\paperwidth},\LenToUnit{0.32in}){\makebox(0,0)[b]{\parbox{\textwidth}{\scriptsize\centering \textcopyright~2026 IEEE. Personal use of this material is permitted. Permission from IEEE must be obtained for all other uses, in any current or future media, including reprinting/republishing this material for advertising or promotional purposes, creating new collective works, for resale or redistribution to servers or lists, or reuse of any copyrighted component of this work in other works.}}}}}

\begin{abstract}
Evaluations of LLM forecasters increasingly measure how forecasts improve as richer evidence is supplied. We identify two confounds in this update-slope metric. First, outcome-aware evidence construction substantially inflates measured responsiveness: on the same models and questions, slopes are roughly four times larger than with resolution-blinded evidence ladders. Second, update slope is a change score with a prior-dependent structural ceiling, so raw comparisons can reflect differences in starting calibration rather than updating behavior. The ceiling defines only a feasible region; whether slope and prior calibration are empirically coupled depends on headroom utilization and varies across our Manifold, ForecastBench, and ACLED pools. Despite this variability, conditioning on prior calibration roughly triples the association between intermediate updating and final Brier score under both blinded and outcome-aware construction. Question identity also explains substantially more variance than model identity across the evaluated pools. We recommend building evidence ladders blind to outcomes, conditioning on prior calibration, and reporting headroom-normalized measures alongside raw slopes.
\end{abstract}


\section{Introduction}

Recent work evaluates LLM forecasters not only by static accuracy \citep{halawi2024approaching,karger2024forecastbench,schoenegger2024wisdom,alur2025aia} but by how forecasts \emph{change} as evidence accumulates \citep{yuan2025assessing}. The natural instrument is an \emph{evidence ladder}: each resolved binary question is presented at increasing context levels, from no context (L0) to a full briefing (L3), and the slope of the truth-aligned forecast against ladder level measures responsiveness to evidence. The natural temptation is to rank models by that slope.

This paper argues that raw update slope is unreliable for two independent reasons, one structural and one methodological, and that both are instances of problems with long histories outside LLM evaluation.

First, the slope is a \emph{change score}, and correlating change with its own baseline is a classical trap: mathematical coupling and regression to the mean can manufacture baseline--change associations without any underlying behavioral difference \citep{oldham1962note,blomqvist1977relation,tu2007revisiting}. In our setting the trap has a sharp finite-range form: because truth-aligned forecasts live in $[0,1]$, the ordinary-least-squares slope obeys a prior-dependent ceiling (\cref{prop:bound}). Crucially, the ceiling licenses only a \emph{feasible region}. Whether observed slopes actually correlate with prior calibration depends on how models use their available headroom, which is an empirical property of a particular panel on a particular pool; on our data it varies in sign across pools.

Second, the evidence ladder itself can leak. If the ladder builder knows the resolution outcome, it can select and strengthen evidence toward the correct side without ever stating it, inflating measured slopes in a way that audits of explicit hindsight markers do not detect. Our primary experiments therefore use \emph{resolution-blinded} ladders; the paired comparison against an earlier outcome-aware construction shows roughly fourfold slope inflation on the same models and questions (\cref{sec:leakage}).

What survives both corrections is the paper's core inferential result: conditioning on prior calibration roughly triples the measured association between intermediate updating and final skill, nearly identically under both constructions (\cref{sec:hier}).

Concretely, this paper contributes:
\begin{itemize}
  \item A tight structural bound on update slope (\cref{prop:bound}), presented as a feasible-region result, with a utilization decomposition $\bar\beta_m = K_3\,u_m\,H_m$ separating what the theorem guarantees from what data must establish.
  \item Classical change-score diagnostics, normalized gain \citep{hake1998interactive}, Oldham's method \citep{oldham1962note}, and a Blomqvist-style error correction \citep{blomqvist1977relation}, applied to LLM dose-response evaluation.
  \item A blinded-ladder design whose paired comparison with outcome-aware construction quantifies the inflation while showing the prior-controlled result is robust to it.
  \item A cross-pool contingency map (Manifold, ForecastBench, ACLED): the ceiling and question-variance dominance replicate everywhere while the slope--prior coupling varies in sign.
\end{itemize}

\section{Related Work}

\paragraph{LLM forecasting and updating}
LLM forecasting pipelines approach crowd performance on contemporary benchmarks \citep{halawi2024approaching,schoenegger2024wisdom,alur2025aia}, though experts remain ahead on rigorous benchmarks \citep{karger2024forecastbench} and evaluation is fragile to simulated ignorance, contamination, and gaming \citep{li2026simulated,paleka2025pitfalls}. Closest to us, Yuan et al.~\citep{yuan2025assessing} find LLMs update conservatively and inconsistently; we show the natural follow-on metric, the update slope, is not a stand-alone measure of updating skill.

\paragraph{Change scores and baseline coupling}
The statistical core of our critique is old. Correlating change with initial value invites mathematical coupling and regression-to-the-mean artifacts; Oldham proposed testing change against the mean of initial and final values \citep{oldham1962note}, Blomqvist derived corrections for measurement error in the baseline \citep{blomqvist1977relation}, and the issue recurs across clinical and educational measurement \citep{tu2007revisiting}. Hake's normalized gain, change divided by maximum possible change, is the standard headroom-normalized effect size in education research \citep{hake1998interactive} and is the endpoint-based analogue of our slope-based utilization quantity $u_m$. Our contribution is to identify how this classical problem manifests in sequential-evidence LLM evaluation, derive the setting-specific finite-range ceiling, and evaluate prior-conditioned alternatives.

\paragraph{Question-level variance}
Murphy~\citep{murphy2026agentic} fits a hierarchical model to per-(forecaster, question) Brier scores on ForecastBench and attributes roughly $62\%$ of variance to question random effects. Our crossed random-effects decompositions on three distinct pools are directionally consistent: question identity dominates model identity in every pool we evaluate, though our seven-model panels leave the model component near the boundary of identifiability. Separately, the reliability--resolution--uncertainty partition of the Brier score that we use for metric construction is due to Allan H.\ Murphy \citep{murphy1973vector}.

\section{Evidence Ladders and the Prior Bound}

\subsection{Blinded Evidence Ladders}

For each resolved binary question $q$ with outcome $y_q \in \{0,1\}$ we construct a four-level ladder $(c_0, c_1, c_2, c_3)$: $c_0$ empty (L0), $c_1$ neutral background (L1), $c_2$ a short factual evidence summary (L2), $c_3$ corroborating facts plus analysis (L3). In our \emph{primary} construction the builder LLM receives the question, resolution criteria, and pre-question background but \emph{not} the outcome; no level may state the resolution. As a sensitivity comparison we retain an earlier \emph{outcome-aware} construction in which the builder saw the outcome but was instructed not to state it. A pre-registered content audit of the outcome-aware ladders (hindsight-marker phrases, resolution-era dates, resolution-blind prediction asymmetry) found only $1.67\%$ aggregate incidence of explicit leakage; \cref{sec:leakage} shows why such audits are insufficient. Ladder prompts, texts, and the audit protocol are in the released repository.

\subsection{Elicitation and Slope}

Given model $m$, question $q$, and level $\ell \in \{0,\dots,L\}$ with $L=3$, we draw $K=3$ probability forecasts with a fixed prompt and average them. The \emph{truth-aligned} forecast is $a_{m,q,\ell} = \bar p_{m,q,\ell}$ if $y_q=1$ and $1-\bar p_{m,q,\ell}$ otherwise; the \emph{update slope} $\beta_{m,q}$ is the OLS coefficient of $a_{m,q,\ell}$ on $\ell$, and $\bar\beta_m$ its per-model mean. Cells missing a parseable probability at some level still contribute a slope fit on the available levels (at least two) to per-model means; analyses requiring all four levels use complete cells only (\cref{sec:setup}).

\subsection{The Prior Bound}

The intuition is elementary. A truth-aligned trajectory starts at $a_{m,q,0}$ and cannot exceed $1$, so the total room to move is $1-a_{m,q,0}$; a slope is a weighted average of per-step moves, so the most extreme slope consistent with the range constraint is proportional to that room. A model that starts well calibrated \emph{cannot} exhibit a large slope, however skilled its updating.

\begin{proposition}[Slope is bounded by prior miscalibration]
\label{prop:bound}
Let $a_{m,q,0}, \ldots, a_{m,q,L} \in [0,1]$ be a monotone non-decreasing truth-aligned trajectory. Set $k = \lceil (L+1)/2 \rceil$. Then the OLS slope satisfies
\[
  \beta_{m,q} \;\leq\; K_L\,\bigl(1 - a_{m,q,0}\bigr),
  \quad K_L = \frac{6\,k\,(L-k+1)}{L(L+1)(L+2)},
\]
with $K_3 = 0.4$. The bound is tight, achieved by the trajectory that stays at $a_{m,q,0}$ for $\ell < k$ and jumps to $1$ for $\ell \ge k$.
\end{proposition}

\begin{proof}[Sketch]
The constant term contributes nothing to the OLS numerator $\sum_\ell (\ell - L/2)\,a_\ell$; under monotonicity with $a_0$ fixed the numerator is maximized by the bang-bang trajectory, and direct computation gives the constant. A weak-monotonicity relaxation ($a_{\ell+1} \ge a_\ell - \epsilon$) adds $C_L\epsilon$ with $C_3 = 0.2$; derivations are in the repository.
\end{proof}

We verify monotonicity where the bound is applied: on both panels, per-model mean trajectories are monotone for four of seven models with maximum violation $0.029$, so the relaxed ceiling exceeds the strict one by at most $C_3\epsilon \approx 0.006$. On the outcome-aware panel all $\bar\beta_m$ lie far below even the strict ceilings; on the blinded panel, where slopes are of the same order as this slack, we treat utilization values as descriptive ratios rather than certified bounds. At the individual-cell level only $\fracMonoCells\%$ of trajectories are weakly monotone (median dip $\cellEpsMedian$, 90th percentile $\cellEpsPninety$), so we do not invoke the bound cell-wise; the hierarchical analysis of \cref{sec:hier} is a separate empirical diagnostic that does not rely on it.

\subsection{Headroom Utilization}
\label{sec:util}

The bound motivates a decomposition rather than a prediction. Define headroom $H_m = 1 - \bar a_{m,0}$ and utilization
\[
  u_m \;=\; \frac{\bar\beta_m}{K_3\,H_m},
\]
so that identically $\bar\beta_m = K_3\,u_m\,H_m$. Under monotonicity \cref{prop:bound} gives $u_m \le 1$ (at most $1 + C_3\epsilon/(K_3 H_m)$ under the relaxation); $u_m$ can be negative when a model moves away from the truth. A positive cross-model correlation between slope and prior miscalibration arises \emph{if utilization varies little across the panel}, in which case slopes are approximately rescaled headrooms; but utilization stability is an empirical property of a model panel on a question pool, not a consequence of the theorem. This distinction, absent from the submitted version of this work and correctly demanded by a reviewer, organizes the empirical sections below: the bound tells us where slopes may live, the data must tell us where they do.

\section{Experimental Setup}
\label{sec:setup}

\paragraph{Questions and models}
The primary pool is $\nQuestions$ resolved binary Manifold Markets questions (volume $\ge 1000$, $\ge 25$ unique bettors, personal and sports markets excluded), all resolving after the latest training cutoff in the panel; two further pools (ForecastBench: RAND INFER and Metaculus; ACLED templated event counts) are used for cross-pool replication with the outcome-aware protocol. The panel is $\nModels$ frontier LLMs: four reasoning (GPT-5, o3, Claude Opus 4.7, GPT-5 mini) and three non-reasoning (GPT-4o, Claude Sonnet 4.6, Claude Haiku 4.5).

\paragraph{Cell accounting}
Of $280$ model--question cells, $\nCellsOrig$ are complete at all four levels in the outcome-aware run: $\nNullRecords$ elicitations returned no parseable probability, leaving $\nCellsIncomplete$ cells (six each for Sonnet~4.6 and Haiku~4.5, almost all at the no-context L0 or L1 rungs) without a valid sample at some level. The blinded run is complete ($\nCellsBlind$ cells). Hierarchical fits use complete cells; the blinded fit restricted to the $268$ cells common to both runs is reported for direct comparability. The missingness is plainly not at random (no-context refusals by two models), so the outcome-aware prior estimates for Sonnet~4.6 and Haiku~4.5 rest on their answered questions only; the complete blinded run is unaffected.

\section{Results}

\subsection{Outcome-Aware Construction Inflates Slopes}
\label{sec:leakage}

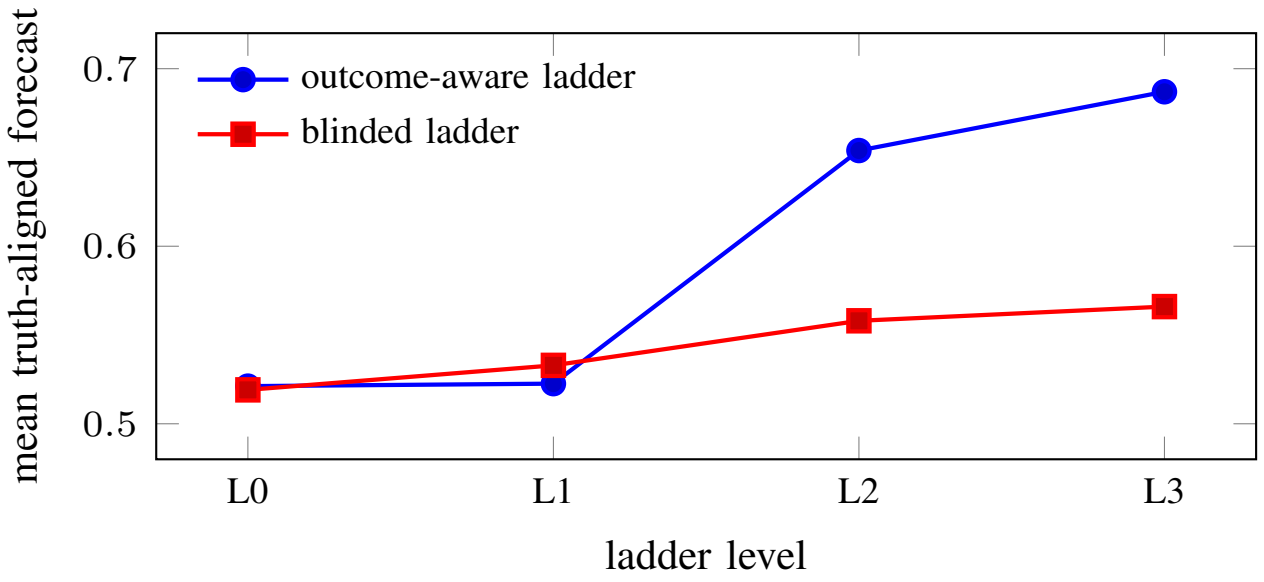
\begin{figure}[t]
  \centering
  \begin{tikzpicture}
    \begin{axis}[
      width=\columnwidth, height=4.4cm,
      xlabel={ladder level}, ylabel={mean truth-aligned forecast},
      xtick={0,1,2,3}, xticklabels={L0,L1,L2,L3},
      ymin=0.48, ymax=0.72,
      legend pos=north west,
    ]
      \addplot+[reasoning, mark=*, thick] coordinates {\aggCoordsOrig};
      \addlegendentry{outcome-aware ladder}
      \addplot+[nonreasoning, mark=square*, thick] coordinates {\aggCoordsBlind};
      \addlegendentry{blinded ladder}
    \end{axis}
  \end{tikzpicture}
  \caption{Aggregate dose response under the two ladder constructions (same models and questions). Starting points nearly coincide at L0, where no ladder content is shown; the outcome-aware ladder then produces roughly $3.5\times$ the aggregate movement of the blinded ladder ($4\times$ in headroom-utilization terms).}
  \label{fig:blind-contrast}
\end{figure}

\begin{table}[t]
  \centering
  \footnotesize
  \setlength{\tabcolsep}{4.5pt}
  \caption{Per-model prior, slope, and utilization $u_m = \bar\beta_m / (K_3 H_m)$ under both constructions ($\bar a_{m,0}$ from the outcome-aware run; blinded $u_m$ uses the blinded run's nearly identical priors).}
  \label{tab:utilization}
  \begin{tabular}{lccccc}
    \toprule
    & & \multicolumn{2}{c}{outcome-aware} & \multicolumn{2}{c}{blinded} \\
    \cmidrule(lr){3-4}\cmidrule(lr){5-6}
    Model & $\bar a_{m,0}$ & $\bar\beta_m$ & $u_m$ & $\bar\beta_m$ & $u_m$ \\
    \midrule
    \uTableRowsBoth
    \bottomrule
  \end{tabular}
\end{table}

\Cref{fig:blind-contrast} and \cref{tab:utilization} present the paired comparison. L0 was independently re-elicited in each run with identical no-context prompts, so the starting points are expected to agree up to sampling noise, and they nearly coincide ($\aggOrigLZero$ vs.\ $\aggBlindLZero$ aggregate aligned forecast). By L3 the outcome-aware ladder has moved the panel to $\aggOrigLThree$ but the blinded ladder only to $\aggBlindLThree$: roughly $72\%$ of the apparent evidence effect disappears when the builder cannot see the outcome. Panel utilization drops from $\uMean$ (range $[\uMin, \uMax]$) to $\uMeanBlind$ (range $[\uMinBlind, \uMaxBlind]$). All blinded slopes remain positive, so the ladder still adds information, but far less than the outcome-aware construction suggested.

Two readings are consistent with this contrast, and we cannot fully separate them: outcome-aware builders may select and strengthen genuinely outcome-consistent evidence (construction leakage), or blinded builders may simply produce weaker evidence for questions whose decisive facts emerged near resolution. Under either reading the methodological conclusion is the same: aggregate dose-response magnitudes are properties of the ladder construction as much as of the models, content-level audits do not bound the effect (our audit found $1.67\%$ explicit leakage against a fourfold slope difference), and evidence ladders for forecasting evaluation should be built blind to outcomes.

\subsection{Prior Control Changes Inferential Conclusions}
\label{sec:hier}

Our primary confirmatory analysis runs on the blinded data. Because the full slope shares $a_{m,q,3}$ with Brier@L3, we use the sub-slope $\beta^{02}_{m,q} = (a_{m,q,2} - a_{m,q,0})/2$, which is out-of-sample for $a_3$, in a crossed mixed-effects model
$\mathrm{Brier@L3} = \alpha + \gamma_1 \beta^{02} + \gamma_2\,\mathrm{Brier@L0} + u_m + u_q + \varepsilon$. The expected sign of $\gamma_1$ is negative: updating toward truth by L2 should predict lower final Brier.

On blinded cells ($n=\nCellsBlind$), the naive coefficient is $\crBlindNaiveBeta$ (CI [$\crBlindNaiveLo$, $\crBlindNaiveHi$]); adding the prior control strengthens it to $\crBlindCtrlBeta$ (CI [$\crBlindCtrlLo$, $\crBlindCtrlHi$]), a roughly threefold increase. The outcome-aware data give nearly identical results ($\crOrigNaiveBeta$ naive, $\crOrigCtrlBeta$ controlled, $n=\nCellsOrig$), and restricting the blinded fit to the $268$ common cells leaves it unchanged ($\crBlindCommonBeta$, CI [$\crBlindCommonLo$, $\crBlindCommonHi$]). The prior-confound correction is therefore robust to the construction leakage of \cref{sec:leakage}: whatever the ladder adds, conditioning on the prior recovers a substantially stronger updating--skill association.

Because $\beta^{02}$ still contains $a_0$, we previously reported an orthogonalized variant as a robustness check; residualizing a predictor on an included control largely reparameterizes the same regression, so we no longer present it as independent evidence. The sharper check uses the L1$\to$L2 update, which contains neither $a_0$ nor $a_3$: controlled for Brier@L0, its coefficient is $\crOrigSlopeTwelveBeta$ (CI [$\crOrigSlopeTwelveLo$, $\crOrigSlopeTwelveHi$]) on outcome-aware data and $\crBlindSlopeTwelveBeta$ (CI [$\crBlindSlopeTwelveLo$, $\crBlindSlopeTwelveHi$]) on blinded data. The association therefore does not depend on shared endpoints; one channel it does not exclude is trajectory persistence, since a forecast that has moved toward truth by L2 tends to remain there at L3. We accordingly read $\gamma_1$ as a predictive association between intermediate progress and final skill under prior control, not as an isolated measure of updating skill.

\subsection{Diagnosing the Slope--Prior Coupling}
\label{sec:mech}

On the outcome-aware Manifold panel, the OLS coefficient of $\bar\beta_m$ on $\sqrt{\mathrm{Brier@L0}}$ is $\couplingSqrtBrier$: worse priors, steeper slopes. \Cref{prop:bound} does not predict this; the bound is far from active (utilization $0.27$--$0.41$). The decomposition of \cref{sec:util} explains it instead. If utilization were exactly constant, the coefficient of slope on headroom $H_m$ would equal $K_3\bar u = \KthreeUMean$. Observed utilization is close to constant (sd $\uSd$, CV $0.16$) but co-varies positively with headroom ($\corrUH$), which pushes the headroom coefficient to $\couplingH$, above $K_3\bar u$. The $\sqrt{\mathrm{Brier}}$ regressor is the root-mean-square rather than the mean of $1 - a_0$ and is not directly commensurable with $H_m$, so we present the decomposition rather than any single coefficient coincidence. The upshot stands regardless: the coupling is what near-constant utilization produces on this panel; it is a fact about the panel, not about the theorem, and it need not transfer. On the blinded panel utilization is neither large nor stable (mean $\uMeanBlind$, CV $0.47$, utilization--headroom correlation $\corrUHBlind$), and the coupling arises through that correlation instead; across pools it changes sign entirely (\cref{sec:pools}).

Classical diagnostics agree with this reading, treated as descriptive at $n=\nModels$. The correlation of total change with baseline is $\corrChangeBaseline$; under Oldham's method, change against the mean of endpoints, it collapses to $\corrChangeOldham$, the signature of baseline coupling rather than a substantive association. A Blomqvist-style correction using the $K{=}3$ within-cell samples shows measurement error contributes little (error share $\sigmaRatioPct\%$ of baseline variance; slope $\blomqvistObs$ observed vs.\ $\blomqvistCorr$ corrected), ruling out error-driven regression to the mean.

\subsection{Question Identity Dominates Model Identity}

A crossed random-effects decomposition of per-cell Brier@L3 attributes $\creFracQuestionPct\%$ of variance to question identity and $\creFracModelPct\%$ to model identity on the primary panel, with the question share robust to leave-one-model-out re-estimation ([$\looFracQMin\%$, $\looFracQMax\%$]); the same qualitative dominance holds on all three pools. Because the model component sits near the boundary of identifiability with seven models, we read this as evidence that question composition can strongly influence measured forecasting performance, and that benchmarks should report question-stratified rather than aggregate scores, not as a universal statement about forecasting leaderboards.

\subsection{Prior-Conditioned Metrics}
\label{sec:metrics}

\begin{table}[t]
  \centering
  \footnotesize
  \setlength{\tabcolsep}{4pt}
  \caption{Kendall $\tau$ between candidate metrics and the Brier@L3 skill ranking (best $=1$ convention; full-precision scores). Descriptive at $n=\nModels$.}
  \label{tab:alternatives}
  \begin{tabular}{lcc}
    \toprule
    Metric & Kendall $\tau$ & $p$ \\
    \midrule
    Raw slope $\bar\beta_m$                & $+\rankTauBrierlThreeVsslope$      & $\rankPBrierlThreeVsslope$ \\
    Hake normalized gain                   & $\rankTauHake$                     & $\rankPHake$ \\
    $\mathrm{DRR}^{\beta}_m$ (slope residual)      & $+\rankTauBrierlThreeVsdrrslope$   & $\rankPBrierlThreeVsdrrslope$ \\
    Murphy resolution gain                 & $\rankTauMurphyRes$  & $\rankPMurphyRes$ \\
    $\mathrm{DRR}^{\rho}_m$ (resolution residual)  & $+\rankTauBrierlThreeVsdrrres$     & $\rankPBrierlThreeVsdrrres$ \\
    \bottomrule
  \end{tabular}
\end{table}

All rankings in this subsection use the outcome-aware panel, whose slopes carry the construction caveat of \cref{sec:leakage}. We attempted the same rankings on the blinded panel and report the outcome as a finding: with blinded slopes this small ($0.003$--$0.025$), no slope-derived metric carries ranking signal (raw slope $\tau = -0.43$, $\mathrm{DRR}^{\rho}$ $\tau = +0.05$, all $p \ge 0.24$); once construction inflation is removed, the dose-response signal on this pool is too weak to differentiate seven models at all. The outcome-aware rankings below are therefore best read as comparisons under the older, more informative but leak-prone instrument. Raw slope ranks models against final skill at only $\tau = +\rankTauBrierlThreeVsslope$. Prior-conditioned alternatives do better (\cref{tab:alternatives}). The dose-response residual $\mathrm{DRR}^{\rho}_m$ residualizes the Murphy resolution gain, the L0$\to$L3 increase in the resolution component of the Brier partition \citep{murphy1973vector} computed on five equal-width probability bins, against $\sqrt{\mathrm{Brier@L0}}$ across the panel; $\mathrm{DRR}^{\beta}_m$ residualizes raw slope the same way. $\mathrm{DRR}^{\rho}_m$ leads at $\tau = +\rankTauBrierlThreeVsdrrres$, with the un-residualized resolution gain and the Hake-style utilization score in between. With seven models these orderings are exploratory: we present $\mathrm{DRR}^{\rho}$ as a promising prior-conditioned option that performed well on this panel, not as a validated default, and the methodology, diagnose the coupling and condition on the prior where baseline dependence is empirically detected, is the contribution rather than any specific ranking. Rankings use full-precision Brier scores; GPT-4o and Haiku~4.5, which round to the same displayed Brier@L3, differ at full precision ($\brierLThreeGptFourOFull$ vs.\ $\brierLThreeHaikuFull$).

\subsection{Cross-Pool Contingency}
\label{sec:pools}

\begin{table}[t]
  \centering
  \footnotesize
  \setlength{\tabcolsep}{4pt}
  \caption{Cross-pool comparison (outcome-aware protocol; $n{=}7$ models, $n_q{=}40$ per pool).}
  \label{tab:cross-pool}
  \begin{tabular}{lccc}
    \toprule
    & Manifold & ForecastBench & ACLED \\
    \midrule
    OLS $\bar\beta_m$ on $\sqrt{\mathrm{Br@L0}}$ & $\couplingSqrtBrier$ & $\primarySlopeOnSqrtBrL$ & $\acledSlopeOnSqrtBrL$ \\
    $r(\mathrm{Br@L0}, \mathrm{Br@L3})$          & $\manifoldRBrLZeroLThree$ & $\primaryRBrLZeroLThree$ & $\acledRBrLZeroLThree$ \\
    Question variance share                      & $\manifoldQVarPct\%$ & $\primaryQVarPct\%$ & $\acledQVarPct\%$ \\
    Model variance share                         & $\manifoldMVarPct\%$ & $\primaryMVarPct\%$ & $\acledMVarPct\%$ \\
    \bottomrule
  \end{tabular}
\end{table}

Running the identical protocol on two further pools (\cref{tab:cross-pool}), with predictions registered in advance, three findings replicate universally: all observed per-model slopes lie below the strict numerical ceiling, every model updates positively, and question variance dominates. The slope--prior coupling does not replicate: $\couplingSqrtBrier$ on Manifold, $\primarySlopeOnSqrtBrL$ on ForecastBench, $\acledSlopeOnSqrtBrL$ on ACLED. Three of nine pre-registered predictions failed, all in this coupling family. The utilization decomposition offers a post-hoc account consistent with these failures: the coupling requires utilization stability, which no theorem supplies and which evidently varies across pools and constructions; with seven models per pool we cannot rule out sampling variability as a contributor, and we present the pool-level explanations as hypotheses. This is the sense of the title: the bound is structural, its empirical signature contingent.

\section{Discussion and Limitations}

Three prescriptions follow. First, \emph{construct evidence ladders blind to outcomes}: outcome-aware construction inflated aggregate dose response roughly fourfold here, and explicit-marker audits could not have detected it. Second, \emph{condition on the prior}: raw slope conflates updating skill with headroom, the classical change-score correction applies, and on both of our Manifold datasets prior control roughly triples the estimated updating--skill association. Third, \emph{report utilization}: $u_m$ makes the headroom adjustment explicit, connects to a mature normalized-gain literature, and travels across pools even when raw couplings do not.

Limitations. The panel is seven models and forty questions per pool; every cross-model correlation, ranking, and diagnostic in this paper is descriptive at that size, and the model variance components sit near identifiability boundaries. The blinded and outcome-aware ladders differ in the totality of their evidence, so we cannot apportion the slope gap between builder hindsight and genuinely weaker blinded evidence; both readings, however, support blind construction. Cross-pool runs use the outcome-aware protocol only, so the leakage finding is established on Manifold alone. Ladders were built by a single model family, and residual contamination through pre-cutoff public discussion of effectively settled questions cannot be excluded. Code, raw elicitation data, prompts, audit protocols, and all analysis scripts are available at \url{https://github.com/BenRossJenkins/update-slope-prior-calibration}; every numeric value in this paper regenerates from the committed data without model-API access.

\section{Conclusion}

Raw update slope is doubly untrustworthy as a forecasting-evaluation metric: it is bounded by prior miscalibration, so cross-model comparisons partly measure how wrong models started, and it is substantially inflated by outcome-aware evidence construction, an effect invisible to content audits. What is robust is the correction: conditioning on prior calibration substantially strengthens the measured association between updating and final skill, identically under blinded and outcome-aware evidence. Evaluate updating the way the change-score literature learned to long ago: blind the instrument, condition on the baseline, and normalize by the room to move.

\bibliographystyle{IEEEtran}
\bibliography{paper}

\end{document}